\documentclass[twoside,leqno,twocolumn]{article}

\usepackage[letterpaper]{geometry}

\usepackage{siamproceedings}

\usepackage[T1]{fontenc}
\usepackage{amsfonts}
\usepackage{graphicx}
\usepackage{epstopdf}
\usepackage{enumitem}
\usepackage{hyperref}
\usepackage{url}
\usepackage{amsmath,amsfonts}
\usepackage{textcomp}
\usepackage{xcolor}
\definecolor{mygreen}{RGB}{0,153,76}
\usepackage[utf8]{inputenc} 
\usepackage[T1]{fontenc}    
\usepackage{hyperref}       
\usepackage{url}            
\usepackage{nicefrac}       
\usepackage{microtype}      
\usepackage{xcolor}         
\usepackage{mathtools}    
\usepackage{bm}      
\usepackage{float}
\newif\ifworkinprogress
\workinprogresstrue 

\ifworkinprogress

\else

\fi
\usepackage{enumitem}
\usepackage{graphicx}
\usepackage{graphicx}   
\usepackage{wrapfig}           
\usepackage{multirow}
\usepackage{booktabs}
\usepackage{arydshln}
\usepackage{xcolor}
\usepackage{placeins}

\usepackage{caption}
\usepackage{subcaption}
\ifpdf
  \DeclareGraphicsExtensions{.eps,.pdf,.png,.jpg}
\else
  \DeclareGraphicsExtensions{.eps}
\fi

\newsiamremark{remark}{Remark}
\newsiamremark{hypothesis}{Hypothesis}
\crefname{hypothesis}{Hypothesis}{Hypotheses}
\newsiamthm{claim}{Claim}

\usepackage{amsopn}

\let\algorithm*\relax
\let\algorithm\relax

\expandafter\let\csname algorithm*\endcsname\relax
\expandafter\let\csname endalgorithm*\endcsname\relax
\usepackage[vlined,linesnumbered,ruled]{algorithm2e}
\usepackage{comment}
\begin{document}

\title{\textbf{Dynamic Sheaf Diffusion Networks with Adaptive Local Structure for Heterogeneous Spatio-Temporal Graph Learning}}

\author{
  Abeer Mostafa\textsuperscript{1}, 
  Raneen Younis\textsuperscript{1, 2} and 
  Zahra Ahmadi\textsuperscript{1, 2}
\\
\textsuperscript{1}\small Peter L. Reichertz Institute for Medical Informatics, Hannover Medical School, Hannover, Germany\\
\textsuperscript{2}\small Lower Saxony Center for Artificial Intelligence and Causal Methods in Medicine (CAIMed), Hannover, Germany
}
\date{}
\maketitle






\begin{abstract}
Spatio-temporal processes often exhibit highly heterogeneous and non-intuitive responses to localized disruptions, limiting the effectiveness of conventional message passing approaches in modeling local heterogeneity. We reformulate spatio-temporal forecasting as the problem of learning information flow over locally structured spaces, rather than propagating globally aligned node representations. To this end, we introduce a spatio-temporal sheaf diffusion graph neural network (ST-Sheaf GNN) that embeds graph topology into sheaf-based vector spaces connected by learned linear restriction maps. Unlike prior approaches relying on static or globally shared transformations, our model learns dynamic restriction maps that evolve over time and adapt to local spatio-temporal patterns, enabling more expressive interactions. The proposed framework both theoretically guarantees and empirically demonstrates evidence that the proposed diffusion mechanism mitigates oversmoothing, preserving discriminative node representations even with increasing diffusion layer depth. Experiments on diverse real-world spatio-temporal forecasting benchmarks across multiple domains demonstrate state-of-the-art performance, highlighting the effectiveness of sheaf topological representations as a principled foundation for spatio-temporal graph learning. The code is available at: \url{https://anonymous.4open.science/r/ST-SheafGNN-6523/}.
\end{abstract}

\section{Introduction}

\begin{figure}
    \centering
    \includegraphics[width=\linewidth]{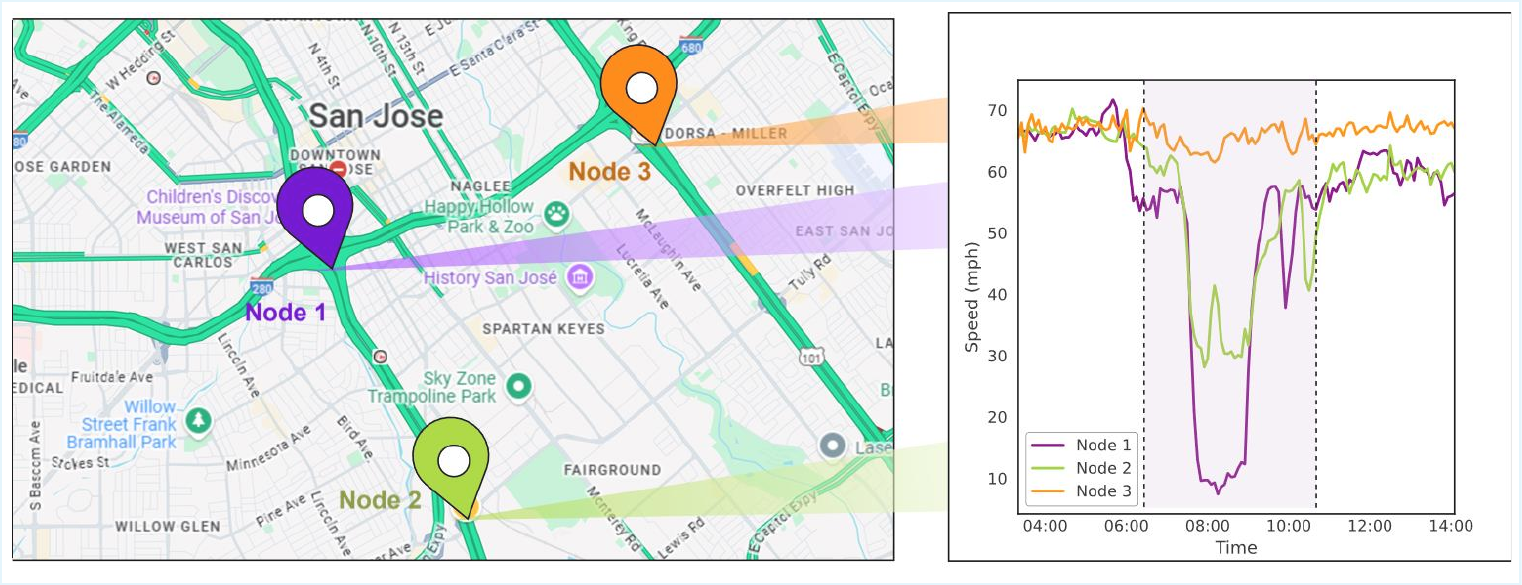}
    \caption{Spatio-temporal traffic dynamics across three sensor nodes in PEMS dataset \cite{li2018diffusion}. Node 2 exhibits a strong temporal correlation with the significant speed drop at Node 1, whereas Node 3, despite similar spatial proximity, remains unaffected.}
\vspace{-5mm}
    \label{intro}
\end{figure}

\leavevmode\\
Spatio-temporal forecasting on graphs is a fundamental challenge in urban computing, environmental monitoring, and infrastructure management, where accurate predictions are critical for real-time decision making. Real-world environments exhibit pronounced spatial heterogeneity and non-stationary dependencies that violate the proximity-based assumptions underlying most graph models: different regions evolve under distinct local dynamics, and interaction strengths are governed not solely by spatial proximity, but by context-dependent relationships that vary continuously across both space and time~\cite{zhou2025fine}. A traffic accident during rush hour, for instance, may trigger cascading congestion across distant neighborhoods while nearby streets remain entirely unaffected (Figure~\ref{intro}). This behavior exposes a core limitation of conventional graph-based approaches: uniform message passing along spatial edges fundamentally cannot capture the heterogeneous interactions~\cite{al2023spatio,jin2023spatio}.

Recent spatio-temporal forecasting models combine graph neural networks \cite{sahili2023spatio, wu2020connecting}, attention mechanisms \cite{feng2023spatiotemporal, 10080902}, and diffusion models \cite{yang2024survey}. While effective in many settings, these methods primarily rely on fixed or uniformly weighted graph propagation and emphasize first-order spatial relationships or simplified temporal dependencies \cite{sahili2023spatio}. Such assumptions neglect the fact that interaction strength can vary significantly across space and time, often leading to oversmoothing and limiting the ability of such models to capture higher-order heterogeneous dependencies inherent in complex spatio-temporal data. As a result, existing methods struggle to adaptively model how information propagates across different regions and time scales.

To address these limitations, we propose a novel spatio-temporal forecasting framework grounded in cellular sheaf theory ~\cite{hansen2019toward}, which generalizes graphs by associating vector spaces (stalks) with nodes and edges, and defining learnable linear maps (restriction maps) that control information flow across the graph in a fully differentiable manner. Unlike static adjacency matrices, sheaf-based models enable \emph{region-specific transformations}, allowing information to propagate differently across the graph (via learnable restriction maps), naturally preserving fine-grained spatial heterogeneity and mitigating oversmoothing~\cite{bodnar2022neural, hansen2021opinion}. The induced sheaf Laplacian generalizes the classical graph Laplacian to support higher-order interactions to be encoded through the underlying cellular structure.
Building on this foundation, we further introduce \emph{dynamic restriction maps} that evolve with spatio-temporal context. This design reframes spatio-temporal forecasting from aggregating neighbor features using fixed propagation rules to learning data-driven geometric operators that capture how local states co-evolve across the graph over space and time. 

We evaluate our framework on six widely used benchmarks (METR-LA, PEMS04, PEMS08, NAVER-Seoul, Molene, AirQuality) across multiple prediction horizons and environmental settings. Experimental results demonstrate consistent improvements over state-of-the-art methods, highlighting the effectiveness and scalability of our approach for real-world applications.
This work makes the following contributions:
\begin{itemize}[leftmargin=*,nosep]
\item We propose the first dynamic sheaf-based formulation for spatio-temporal learning, modeling graph topology via learned, locally heterogeneous restriction maps, that unifies heterogeneous spatial and temporal dynamics and effectively mitigates oversmoothing in deep GNN architectures.
\item We design a \emph{dynamic sheaf diffusion operator} that captures heterogeneous spatio-temporal interactions while remaining efficient and scalable. 
\item Extensive experiments across multiple domains demonstrate state-of-the-art performance and substantially improved expressive power compared to existing spatio-temporal graphical models.
\end{itemize}

\section{Related Work}
\label{sec:related_work}
\subsection{Spatio-Temporal Forecasting:} 
Spatio-temporal forecasting on graphs has progressed through several architectural generations. Early methods combined graph convolutions with sequential models to jointly capture spatial and temporal structure: STGCN~\cite{yu2018spatio} integrated spatial graph convolutions with temporal 1D convolutions in a unified feedforward framework, while DCRNN~\cite{li2018diffusion} modeled bidirectional diffusion on directed graphs via gated recurrent units. Recognizing the rigidity of fixed adjacency matrices, subsequent work introduced data-driven graph learning: Graph WaveNet~\cite{wu2019graph} learns latent spatial dependencies entirely from data through adaptive node embeddings, freeing the model from reliance on pre-defined graph structure. Attention mechanisms further enhanced this direction, with ASTGCN~\cite{guo2019attention} incorporating joint spatial and temporal attention for context-aware feature weighting, and GMAN~\cite{zheng2020gman} introducing spatio-temporal positional encodings within a multi-attention framework to capture long-range interaction patterns. More recent efforts have focused on scalability and spectral expressivity: SGP~\cite{cini2023scalable} achieves computational efficiency through randomized recurrent architectures with lightweight spatial encodings, CITRUS~\cite{einizade2024continuous} exploits the separability of continuous heat kernels on Cartesian graph products for multi-scale spectral decomposition, and STDN~\cite{cao2025spatiotemporal} disentangles trend and seasonal components per node to model non-stationary temporal dynamics. Despite these advances, all of these methods remain fundamentally constrained to first-order graph structures that enforce uniform information propagation across edges and assume all nodes reside in a shared latent space; assumptions that are ill-suited for heterogeneous spatio-temporal dynamics and motivate a shift toward topologically richer representations.

\subsection{Sheaf Graph Neural Networks:}
Recent efforts present a move beyond traditional spatial representations to capture higher-order interactions and address oversmoothing limitations in GNNs through sheaf theory~\cite{suk2022surfing, braithwaite2024heterogeneous}. Sheaf neural networks represent a promising direction that applies concepts of algebraic topology to model asymmetric data relationships~\cite{hansen2020sheaf, bodnar2022neural}. Building on cellular sheaf theory, SheafANs \cite{barbero2022sheaf} developed a sheaf attention mechanism that generalizes graph attention networks by integrating cellular sheaves for richer geometric inductive biases. This approach addresses the GNN limitations, oversmoothing, and poor performance on heterophilic graphs by using transport matrices and sheaf-based feature aggregation to preserve local heterogeneity and geometric structure. In another work, cellular sheaves and both linear and non-linear sheaf hypergraph Laplacians are introduced to improve the generalization of standard hypergraph Laplacians \cite{duta2023sheaf}.
Our work is the first study that extends the sheaf theory to model the asymmetric higher-order dependencies for spatio-temporal problems. 

\section{Proposed Framework: ST-Sheaf GNN}

\leavevmode\\
\begin{figure*}[ht!]
    \centering
    \includegraphics[width=0.95\linewidth]{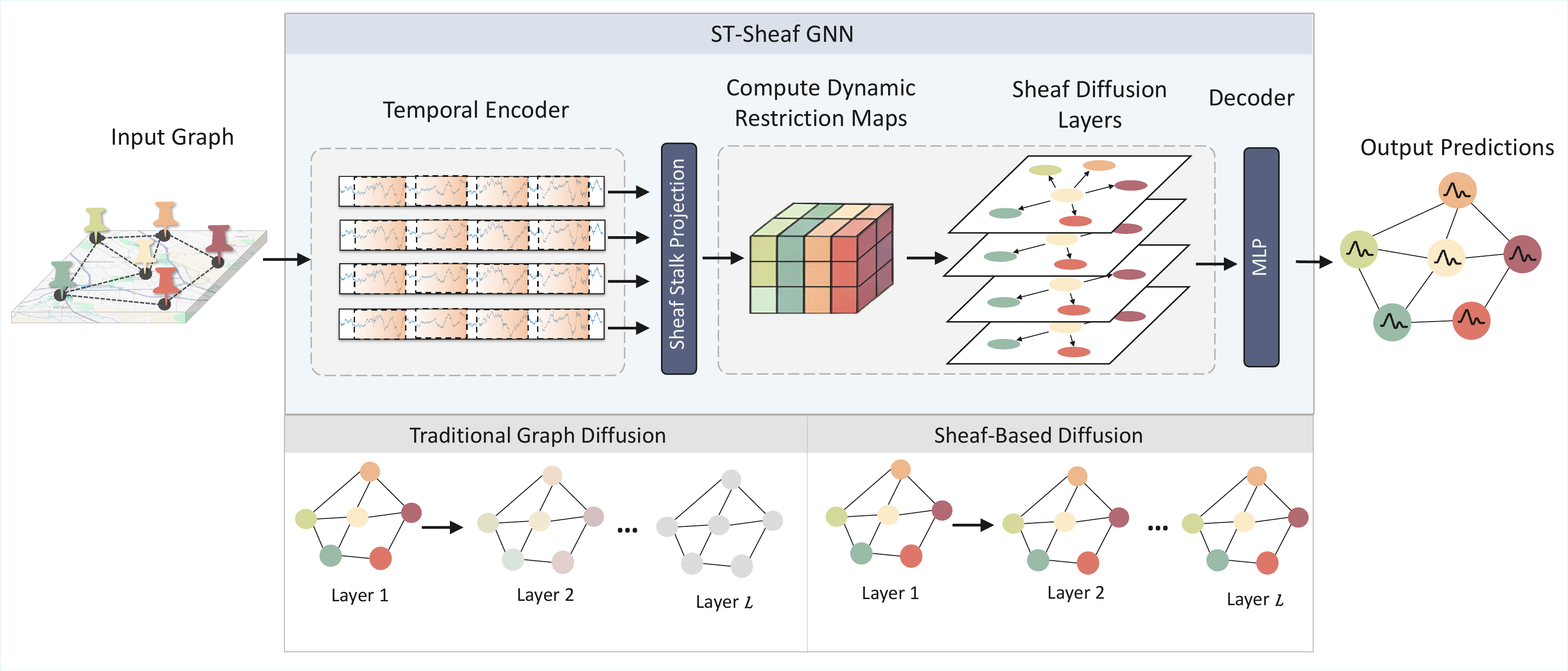}
    \caption{Overview of the proposed ST-Sheaf GNN. Multivariate time-series signals on a graph are first encoded with multi-head attention to capture temporal dependencies. The resulting representations are then projected into a sheaf space, where dynamic restriction maps are computed. Sheaf diffusion layers propagate information by preserving local distinct node representation. The diffused features are finally decoded by an MLP to produce node-level forecasts.}
    \label{fig:methodology}
    \vspace{-3mm}
\end{figure*}
In this section, we present ST-Sheaf GNN, a spatio-temporal forecasting framework that combines temporal self-attention with sheaf-based spatial diffusion to model complex dependencies on graph-structured time series. Given a graph whose nodes are associated with multivariate temporal signals, the model first encodes node-wise temporal dynamics through self-attention over the historical observation window. It then performs spatial reasoning at each time step using a cellular sheaf formulation, in which node representations evolve in heterogeneous local coordinate systems and interact through learned, signal-conditioned restriction maps. This design enables ST-Sheaf GNN to capture asymmetric and higher-order spatial dependencies while mitigating the oversmoothing behavior of conventional message-passing GNNs. The overall architecture is illustrated in Figure~\ref{fig:methodology}. We begin by formalizing the problem, then describe the temporal encoder and the sheaf structure construction, including the sheaf Laplacian and the gated sheaf diffusion mechanism, to generate future predictions.


\subsection{Problem Definition: }
Let $\mathcal{G} = (\mathcal{V}, \mathcal{E})$ denote a graph with $N = |\mathcal{V}|$ nodes and $E = |\mathcal{E}|$ edges. 
We consider a spatio-temporal forecasting problem in which each node is associated with a multivariate time series. 
Given an observed historical window: 
\[
\mathbf{X} \in \mathbb{R}^{B \times T \times N \times F_{\mathrm{in}}},
\]
with batch size $B$, sequence length $T$, and input feature dimension 
$F_{\mathrm{in}}$, the objective is to predict future node-level signals for output feature dimension $F_{\mathrm{out}}$:
\[
\mathbf{Y} \in \mathbb{R}^{B \times H \times N \times F_{\mathrm{out}}},
\]
over a forecasting horizon $H$.

\subsection{Temporal Encoding: }
To capture temporal dependencies, we apply self-attention along the time dimension independently for each node. Let
\[
\mathbf{Z}_v = [\mathbf{z}_{1,v}, \ldots, \mathbf{z}_{T,v}] \in \mathbb{R}^{T \times D},
\]
denote the embedded sequence for node $v$. 
A multi-head self-attention block (MHA) computes:
\begin{equation}
\mathbf{Z}_v^{(1)}
=
\mathrm{MHA}(\mathbf{Z}_v, \mathbf{Z}_v, \mathbf{Z}_v)
+
\mathbf{Z}_v,
\label{eq:temporal_attn}
\end{equation}
where residual connections preserve the input signal. The output is further refined via layer normalization and a position-wise feed-forward network:
\begin{align}
\begin{split}
\mathbf{Z}_v^{(2)} &= \mathrm{LayerNorm}(\mathbf{Z}_v^{(1)}), \\
\mathbf{Z}_v^{(3)} &= \mathrm{FFN}(\mathbf{Z}_v^{(2)}) + \mathbf{Z}_v^{(2)}.
\end{split}
\label{eq:temporal_ffn}
\end{align}

Finally, the temporally encoded features are projected to a latent space that interfaces with subsequent spatial modeling:
\begin{equation}
\mathbf{h}_{t,v}
=
\mathbf{W}_{\mathrm{proj}} \mathbf{z}_{t,v}^{(3)},
\qquad
\mathbf{h}_{t,v} \in \mathbb{R}^{d},
\label{eq:stalk_projection}
\end{equation}
where $\mathbf{W}_{\mathrm{proj}} \in \mathbb{R}^{d \times D}$ is a learnable projection matrix. Sheaf diffusion is applied independently at each time step $t \in \{1, \ldots, T\}$; when the time index is clear from context, we write $\mathbf{h}_u = \mathbf{h}_{t,u}$ for simplicity.

\subsection{Sheaf Building: }
Sheaf theory provides a principled way to model heterogeneous local coordinate systems on graphs. A cellular sheaf $\mathcal{F}$ over $\mathcal{G}$ assigns:
\begin{itemize}[nosep]
    \item a \emph{vertex stalk} $\mathcal{F}(v) = \mathbb{R}^d$ to each node $v \in \mathcal{V}$,
    \item an \emph{edge stalk} $\mathcal{F}(e) = \mathbb{R}^d$ to each edge $e \in \mathcal{E}$,
    \item linear \emph{restriction maps}
    \[
    \rho_{v \trianglelefteq e} : \mathcal{F}(v) \rightarrow \mathcal{F}(e),
    \quad \forall\, v \in e.
    \]
\end{itemize}
\begin{figure}[t!]
    \centering
    \includegraphics[width=0.78\linewidth]{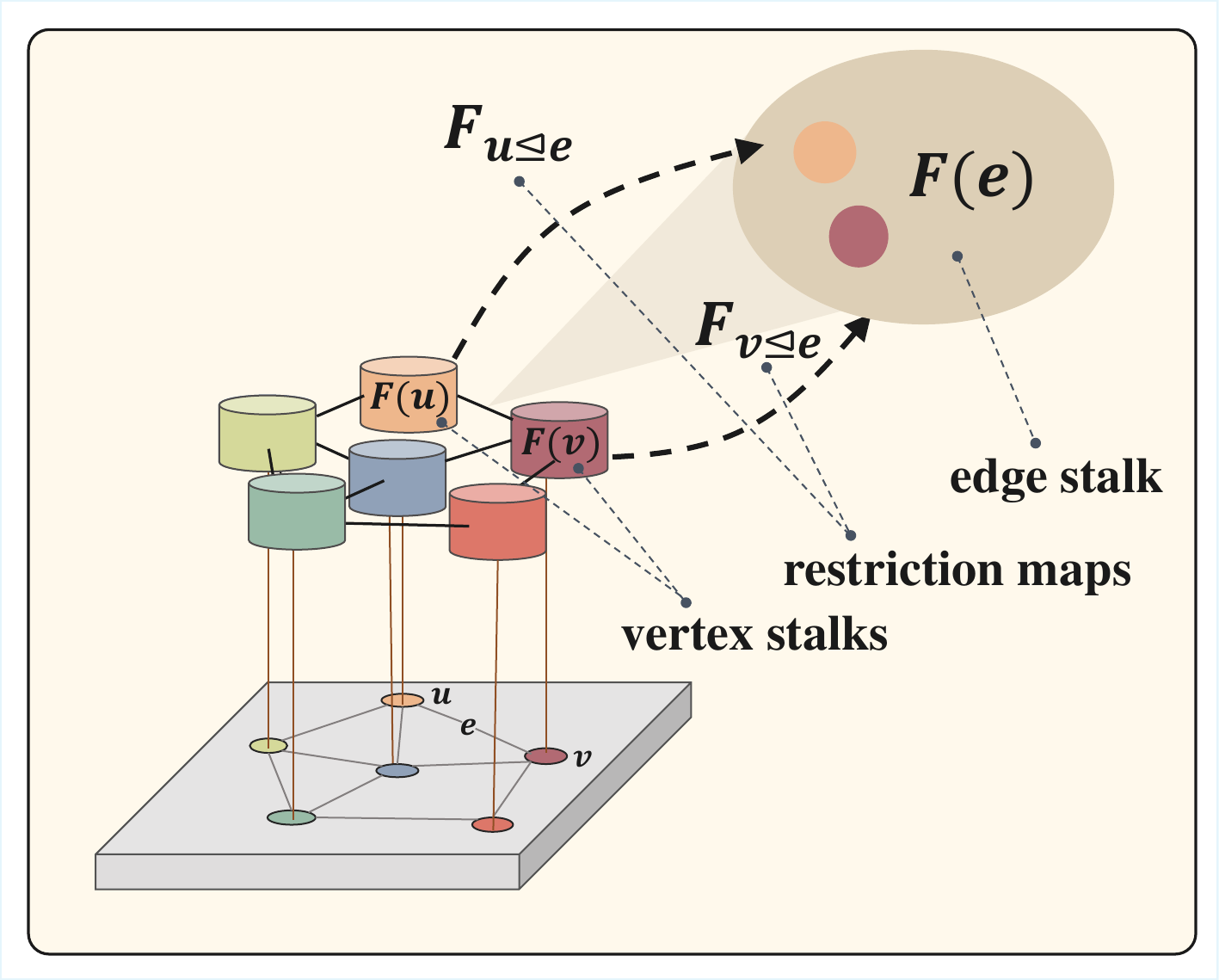}
    \caption{Sheaf graph representation with learned dynamic restriction maps. Each node is associated with a vertex stalk, and each edge is associated with an edge stalk. Restriction maps define the relation between vertex stalks and edge stalks.}
\vspace{-3mm}
    \label{fig:sheaf}
\end{figure}

Each vertex stalk defines a local latent coordinate system in which node-level signals are represented. Unlike standard GNNs, neighboring nodes are not required to share a common representation space.
Restriction maps align node representations to an edge-local space, enabling meaningful interaction between neighboring nodes with potentially different local coordinates. Figure~\ref{fig:sheaf} illustrates the resulting sheaf-based topology.

A key hyperparameter is the stalk dimension $d$, which determines the capacity of the local vector spaces. Higher-dimensional stalks allow richer feature propagation and more expressive signal representations. When $d=1$ and $\rho_{v \trianglelefteq e}$ is identity, the formulation reduces to scalar-weighted message passing as in classical graph convolution networks (GCNs). 
In our model, we assume a uniform stalk dimension so that: 
\begin{equation}
    \mathcal{F}(v) \cong \mathcal{F}(e) \cong \mathbb{R}^d \qquad
\forall  v \in \mathcal{V}, e \in \mathcal{E}. 
\end{equation}

For an edge $e = (u,v)$, vertex-level representations are first aligned to the corresponding edge stalk via the restriction maps, yielding $\rho_{u \trianglelefteq e}\mathbf{h}_u$ and $\rho_{v \trianglelefteq e}\mathbf{h}_v$.
These aligned representations lie in a shared edge-local coordinate system and can therefore be compared meaningfully, even if the original node features reside in different latent spaces.
This construction leads to a diffusion mechanism that differs fundamentally from standard GNN message passing, to minimize the Dirichlet energy implicitly: 
\begin{equation}
    \mathcal{E}(\mathbf{h}) = \frac{1}{2} \sum_{(u,v) \in \mathcal{E}} \|\mathbf{h}_u - \mathbf{h}_v\|_2^2.
\end{equation}
Repeated minimization of this energy is a primary cause of oversmoothing: by enforcing global representational consistency regardless of local structure or semantic context, the model progressively erases the very heterogeneity it needs to distinguish nodes, ultimately collapsing all representations within a connected component toward a single fixed point.
In contrast, the sheaf-based energy relaxes this constraint by introducing learnable, edge-specific coordinate transformations. Rather than requiring direct agreement between $\mathbf{h}_u$ and $\mathbf{h}_v$, we penalize disagreement only after aligning representations through restriction maps. This yields a more expressive energy function:
\begin{equation}
    \mathcal{E}_{\mathcal{F}}(\mathbf{h}) = \frac{1}{2} \sum_{e=(u,v) \in \mathcal{E}} \left\| \rho_{u \trianglelefteq e} \mathbf{h}_u - \rho_{v \trianglelefteq e} \mathbf{h}_v \right\|_2^2,
\end{equation}
where $\rho$ is the linear restriction map that projects the vertex representation into the edge-local coordinate. The key insight is that these transformations break the uniform aggregation pattern of standard GNNs, allowing neighboring nodes to preserve distinct local representations even after multiple diffusion steps.

\subsection{Sheaf Laplacian:} The gradient of $\mathcal{E}_{\mathcal{F}}$ with respect to $\mathbf{h}$ defines the sheaf Laplacian operator $\mathcal{L}_{\mathcal{F}}$. For a node $u$, this operator aggregates edge-aligned differences across all incident edges:
\begin{equation}
    (\mathcal{L}_{\mathcal{F}} \mathbf{h})_u = \sum_{e: u \in e} \rho_{u \trianglelefteq e}^{\top} \left( \rho_{u \trianglelefteq e} \mathbf{h}_u - \rho_{v \trianglelefteq e} \mathbf{h}_v \right),
\end{equation}
where the sum is taken over all edges $e = (u,v)$ incident to $u$. This operator generalizes the standard graph Laplacian to vector-valued signals defined in heterogeneous local coordinate systems. The restriction maps act as learnable, edge-specific transformations that modulate information flow according to local graph structure and node states. The generalization to higher-dimensional stalks and learnable restriction maps significantly expands the representational capacity of the diffusion process.

A central modeling choice is how to parameterize the restriction maps. In principle, these could be arbitrary linear operators in $\mathbb{R}^{d \times d}$, but this would require $O(Ed^2)$ parameters, and incur substantial computational overhead and overfitting risk, particularly for large graphs.
We instead adopt a diagonal parameterization where each restriction map is represented as element-wise multiplication by a learned vector:
\begin{equation}
    \rho_{u \trianglelefteq e} \mathbf{h}_u = \mathbf{r}_{u \trianglelefteq e} \odot \mathbf{h}_u,
\end{equation}
where $\mathbf{r}_{u \trianglelefteq e} \in \mathbb{R}^d$ is a learnable vector and $\odot$ denotes element-wise multiplication. This reduces the parameter complexity to $O(Ed)$. Crucially, this efficiency gain does not sacrifice expressivity: the diagonal restriction maps still break the uniform aggregation pattern of standard GNNs, and as shown in Appendix \ref{app:proof}, the resulting sheaf Laplacian provably supports heterogeneous equilibrium representations that prevent oversmoothing by construction.

Restriction maps can be learned either statically, computed once from fixed node embeddings and reused across all inputs, or dynamically, conditioned on the current node states. Since edge semantics in spatio-temporal data often vary over time, we adopt dynamic signal-conditioned restriction maps to be learned in the following way:
\begin{equation}
    [\mathbf{r}_{u \trianglelefteq e}, \mathbf{r}_{v \trianglelefteq e}] = \text{MLP}(\mathbf{h}_u \| \mathbf{h}_v]).
\end{equation}
This input-conditioned design is a critical distinction from prior sheaf-based models that rely on static, pre-computed restriction maps: by conditioning on current node states, the sheaf geometry itself evolves dynamically, allowing the model to capture time-varying edge semantics that static maps fundamentally cannot represent. To stabilize training, we further incorporate a residual connection in restriction map computations with a small initialization:
\begin{equation}
    \mathbf{r}_{u \trianglelefteq e} \leftarrow \mathbf{r}_{u \trianglelefteq e} + \alpha \cdot \text{MLP}_{\text{res}}([\mathbf{h}_u \| \mathbf{h}_v]),
\end{equation}
where $\alpha$ is a scaling coefficient and $\text{MLP}_{\text{res}}$ is initialized with near-zero weights.
This is empirically validated in our ablation studies, where replacing dynamic maps with static alternatives leads to consistent performance degradation across all datasets, confirming that time-varying edge semantics are a critical factor in spatio-temporal forecasting.

To avoid domination by high-degree nodes during diffusion, we apply degree-based normalization. For each edge $e = (u,v)$, we define:
\begin{equation}
    w_e = \frac{1}{\sqrt{\deg(u) \cdot \deg(v)}},
\end{equation}
where $\deg(u)$ denote the degree of node $u$.

\subsection{Sheaf Diffusion:}
Sheaf diffusion can be interpreted as heat diffusion over a sheaf-structured graph. It enables information exchange across aligned local spaces while preserving feature heterogeneity, and strengthens the capacity to linearly separate node representations as the diffusion process converges. 
For each edge $e$, we compute the sheaf-aligned discrepancy $\boldsymbol{\delta}_e = (\mathbf{r}_{u \trianglelefteq e} \odot \mathbf{h}_u) - (\mathbf{r}_{v \trianglelefteq e} \odot \mathbf{h}_v)$. Messages are then aggregated using a signed scatter operation:
\begin{equation}
    \mathbf{m}_u = \sum_{e: \mathrm{src}(e)=u} w_e \boldsymbol{\delta}_e - \sum_{e: \mathrm{dst}(e)=u} w_e \boldsymbol{\delta}_e.
    \label{eq:message_agg}
\end{equation}
The aggregated message $\mathbf{m}_u$ is transformed and combined with the original representation $\mathbf{h}_u^{(\ell)}$ through a residual connection and feed-forward block to produce a candidate update $\tilde{\mathbf{h}}_u$. A learnable gate then adaptively blends the original and updated representations:

\begin{align}
\begin{split}
    \mathbf{g}_u &= \sigma\big(\mathbf{W}_g^{(\ell)}[\mathbf{h}_u^{(\ell)} \,\|\, \tilde{\mathbf{h}}_u]\big), \\
    \mathbf{h}_u^{(\ell+1)} &= \mathbf{g}_u \odot \tilde{\mathbf{h}}_u + (1 - \mathbf{g}_u) \odot \mathbf{h}_u^{(\ell)}.
\label{eq:gated_update}
\end{split}
\end{align}    

While residual connections stabilize gradient flow in deep architectures, the gating mechanism provides feature-wise control over the trade-off between preserving existing information and incorporating diffusion-driven updates. When $\mathbf{g}_u \approx 0$, the model preserves the current representation; when $\mathbf{g}_u \approx 1$, it favors diffusion-driven updates. 
After $L$ diffusion layers, the final representations are decoded through a linear projection to produce forecasts $\hat{\mathbf{Y}} \in \mathbb{R}^{B \times H \times N \times F_{\mathrm{out}}}$.

\section{Experimental Setup} 
\label{sec:exp}

\leavevmode\\
We evaluate ST-Sheaf GNN on six real-world benchmark datasets spanning traffic, weather, and air quality domains. For traffic forecasting, we use METR-LA~\cite{li2018diffusion}, a freeway sensor dataset from Los Angeles; PEMS04 and PEMS08~\cite{song2020spatial}, two regional subsets of the Caltrans Performance Measurement System (PeMS) with distinct graph topologies and flow dynamics; and NAVER-Seoul~\cite{leelearning}, a large-scale urban dataset covering the entire road network of Seoul, South Korea, characterized by abrupt speed fluctuations and a substantially larger sensor count, making it a particularly demanding benchmark for scalability and generalization. To assess cross-domain generalization, we additionally evaluate on the Molene weather dataset~\cite{girault2015stationary}, comprising hourly meteorological measurements from the Brest region of France, and the Air Quality dataset~\cite{urbanair_microsoft_research}, which contains hourly $\text{PM}_{2.5}$ concentration readings from 437 monitoring stations across 43 Chinese cities. Table~\ref{tab:dataset_summary} provides the summary of the characteristics of all datasets.

We follow established data splitting protocols: PEMS04 and PEMS08 are split using a 60:20:20 train/validation/test ratio ~\cite{song2020spatial, gao2024spatial, wang2025dual}, while METR-LA, NAVER-Seoul, Molene, and AirQuality follow a 70:10:20 split~\cite{leelearning, li2018diffusion, jiang2023spatio}. Graph structures are constructed from pre-defined spatial adjacency matrices derived from physical distances between sensor locations.
All experiments are conducted in PyTorch on an NVIDIA GeForce RTX 4070 GPU (8 GB). We train using the Adam optimizer with an initial learning rate of $0.01$, a mini-batch size of 12, and MAE as the training loss. Early stopping is applied based on validation loss with a patience of 10 epochs. Each experiment is repeated three times, and the results are averaged. Model hyperparameters are selected empirically per dataset: the stalk dimension $d$ is set to 16 for smaller datasets and 32 for larger ones, the number of sheaf diffusion layers ranges from 2 to 4, and the number of temporal attention heads is chosen from $\{4, 8\}$. Complete hyperparameter configurations for all experiments are provided in the code repository.
We report MAE, RMSE, and MAPE for traffic datasets, consistent with prior work, and MAE for weather and air quality datasets. Missing values are excluded from all metric computations, following the same procedure as the baseline methods.
Each model takes as input a fixed historical window and predicts future node-level signals. For traffic datasets (METR-LA, NAVER-Seoul, PEMS04, PEMS08), the input consists of the previous 12 time steps (one hour), with forecasting horizons of 15, 30, and 60 minutes (horizons 3, 6, and 12), following established benchmarking protocols. For the Molene dataset, the model is conditioned on the past 10 hours and evaluated on the next 5 time steps. For the Air Quality dataset, the input spans 48 hours (2 days), and predictions cover the next 48 hours, reported as averages over three sub-horizons: 1-12~h, 13-24~h, and 25-48~h. All inputs are normalized using z-score normalization applied independently per sensor node and inverted prior to evaluation, consistent with the preprocessing adopted by all baseline methods.

We compare ST-Sheaf GNN against a comprehensive set of state-of-the-art spatio-temporal forecasting baselines spanning all major methodological categories outlined in Section~\ref{sec:related_work}. These include classical sequence modeling (ARIMA~\cite{li2018diffusion}), graph convolutional networks (STGCN~\cite{yu2018spatio}, DCRNN~\cite{li2018diffusion}), adaptive graph learning (Graph WaveNet~\cite{wu2019graph}), attention-based models (ASTGCN~\cite{guo2019attention}, GMAN~\cite{zheng2020gman}), scalable recurrent architectures (SGP~\cite{cini2023scalable}), spectral methods (CITRUS~\cite{einizade2024continuous}), and decomposition-based approaches (STDN~\cite{cao2025spatiotemporal}).

\section{Results and Discussion}

\begin{table*}[ht!]
\centering
\small
\caption{Performance on NAVER-Seoul Dataset and METR-LA dataset.}
\label{tab:res1}
\resizebox{0.85\linewidth}{!}{%
\begin{tabular}{l@{\hspace{7pt}}l@{\hspace{8pt}}ccccccccc}
\toprule
\multirow{2}{*}{\textbf{Data}} & \multirow{2}{*}{\textbf{Model}} & \multicolumn{3}{c}{\textbf{15min / horizon 3}} & \multicolumn{3}{c}{\textbf{30min / horizon 6}} & \multicolumn{3}{c}{\textbf{60min / horizon 12}} \\ 
\cmidrule{3-11}
& & MAE & RMSE & MAPE & MAE & RMSE & MAPE & MAE & RMSE & MAPE \\
\midrule
\multirow{10}{*}{\textbf{NAVER-Seoul}} & ARIMA \cite{li2018diffusion}& 5.51 & 8.15 & 15.33 &6.11 &9.22 &18.45 &7.85 &12.23& 22.16\\
& STGCN \cite{yu2018spatio}&  4.63 & 6.92 & 14.49 & 5.50 & 8.83 & 17.37 & 6.77 & 10.89 & 20.42  \\
& DCRNN \cite{li2018diffusion}& 4.86 & 7.12 & 15.35 & 5.67 & 8.80 & 18.38 & 6.40 & 10.06 & 21.09  \\
& GW-Net \cite{wu2019graph}& 4.91 & 7.24 & 14.86 & 5.26 & 8.13 & 16.16 & 5.55 &  8.77 &  16.97  \\
& GMAN \cite{zheng2020gman}& 5.20 & 8.32 & 16.98 & 5.35 & 8.67 & 17.47 &  5.48 & 8.94 & 17.89\\
& ASTGCN \cite{guo2019attention}& 5.09 & 7.44 & 16.14 & 5.71 & 8.73 & 18.78 & 6.22 & 9.58 & 20.37  \\
&SGP \cite{cini2023scalable} 
& 4.87& 7.28& 15.63& 5.48& 8.61&18.12&  6.01& 9.64&20.04\\
& CITRUS \cite{einizade2024continuous}& 5.29 &7.80&16.78 &6.17& 9.56&20.20&7.16&11.31&24.10\\
& STDN \cite{cao2025spatiotemporal}& 4.73 &  6.82 &  14.46 &  5.12 &  7.95 &  16.12 & 5.94 &9.02 & 18.01\\
& \textbf{ST-Sheaf (Ours)} & \textbf{4.52}  & \textbf{6.60}  & \textbf{14.23}  
& \textbf{5.01}  & \textbf{\textbf{7.72}}  & \textbf{16.06}  
& \textbf{5.11} & \textbf{8.32} & \textbf{16.91} \\
\cmidrule{1-11}
\multirow{10}{*}{\textbf{METR-LA}} & ARIMA \cite{li2018diffusion}& 3.99 & 8.21 & 9.60 & 5.15 & 10.45 & 12.70 &  6.90 & 13.23 & 17.40 \\
& STGCN \cite{yu2018spatio}& 2.88 & 5.74 & 7.62 & 3.47 & 7.24 & 9.57 & 4.59 & 9.40 & 12.70  \\
& DCRNN \cite{li2018diffusion}& 2.77 & 5.38 & 7.30 & 3.15 & 6.45 & 8.80 & 3.60 & 7.59 & 10.50 \\
& GW-Net \cite{wu2019graph}& 2.69 & 5.15 & 6.90 & 3.07 & 6.22 & 8.37 & 3.53 & 7.37 & 10.01  \\
& GMAN \cite{zheng2020gman}& 2.86 & 5.77 & 7.76 & 3.14 & 6.59 & 8.73 & 3.48 & 7.35 & 10.10 \\
& ASTGCN \cite{guo2019attention}& 3.25 & 6.28 & 9.27 & 3.81 & 7.56 & 11.34 & 3.58 & 7.73 & 10.35  \\
& SGP \cite{cini2023scalable} 
& 2.90 & 5.76 & 7.78 
& 3.33 & 6.86 & 9.47
& 3.90 & 8.22 & 11.8\\
& CITRUS \cite{einizade2024continuous}
&  2.70 &  5.14 &  6.74
&  2.98& 5.90 & \textbf{7.78}
& 3.44 & 6.85 &  9.28 \\
& STDN \cite{cao2025spatiotemporal}&  2.67 & 5.68 & 7.17 & 3.00 & 6.59 & 8.58 &  3.33 & 7.44 & 10.20\\
& \textbf{ST-Sheaf (Ours)}
& \textbf{2.56} & \textbf{5.06} & \textbf{6.71}
& \textbf{2.96} &  \textbf{5.88} &  8.12
& \textbf{3.25} &  \textbf{6.80} & \textbf{9.22} \\
\bottomrule
\end{tabular}%
}
\end{table*}

\begin{table*}
\centering
\small
\caption{Performance on Molene weather, Air Quality in MAE and PEMS04, PEMS08 traffic datasets.}
\label{tab:res2}
\resizebox{0.99\linewidth}{!}{%
\begin{tabular}{lcccccccccccccc}
\toprule
 \multirow{2}{*}{\textbf{Model}} & \multicolumn{5}{c}{\textbf{Molene (weather)}} & \multicolumn{3}{c}{\textbf{Air Quality (PM2.5)}} & \multicolumn{3}{c}{\textbf{PEMS04}} & \multicolumn{3}{c}{\textbf{PEMS08}}\\ 
\cmidrule{2-15}
 & \textit{step-1} & \textit{step-2} & \textit{step-3} & \textit{step-4} & \textit{step-5} 
 & \textit{1-12 h} & \textit{13-24 hr} & \textit{25-48 hr} & \footnotesize MAE & \footnotesize RMSE & \footnotesize MAPE & \footnotesize MAE & \footnotesize RMSE & \footnotesize MAPE\\
\midrule
ARIMA \cite{li2018diffusion} & 0.445 & 0.750 & 1.041 & 1.327 & 1.592 & 30.14 & 38.98 & 44.10& 33.73 & 48.80 & 24.18 & 31.09 & 44.32 & 22.73 \\
STGCN \cite{yu2018spatio}     & 1.066& 1.552 & 1.933&2.225 & 2.698 & 22.77 & 31.59 & 35.59& 22.70 & 35.55 &  14.59 & 18.02 & 27.83 &  11.40\\
DCRNN \cite{li2018diffusion}   &  0.396&  0.635 &  0.907 &  1.164 &  1.419 & 24.28 & 32.95 & 36.11 & 24.70 & 38.12 & 17.12 & 17.86 & 27.83 & 11.45 \\
GW-Net \cite{wu2019graph}      & 0.855& 1.124 & 1.455&1.865 & 2.104 & 22.42 & 30.29 & 34.73  & 25.45 & 39.70 & 17.29 & 19.13 & 31.05 & 12.68 \\
GMAN \cite{zheng2020gman}      & 0.962& 1.424 & 1.882& 2.022& 2.417 & 24.11 & 32.80& 36.25&  20.23 &  32.17 & 17.06 &  16.47 & 25.72 &  10.48\\
ASTGCN \cite{guo2019attention}  & 0.981 & 1.342 & 1.877 &  1.906& 2.125 & 25.51 & 33.98 & 37.42& 22.80 & 35.82 & 16.56 & 18.63 & 28.27 & 13.08 \\
SGP \cite{cini2023scalable}     & 0.522 & 0.848 & 1.148 & 1.399 & 1.650 & 26.42 &33.38 & 35.99&21.90 &34.27 & 16.15& 18.06&27.33&13.55\\
CITRUS \cite{einizade2024continuous}  & 0.497 & 0.757  & 1.048 & 1.332& 1.600 & 27.40 & 32.66 & 35.56 & 22.64 &35.08 & 16.83&17.89&27.81&11.21\\
STDN \cite{cao2025spatiotemporal} & \textbf{0.352} &  0.621 & 0.924 &  1.336 & 1.566 & \textbf{22.17} &  30.16 &  34.62& \textbf{19.15} & 34.37 & 23.51 &18.26& 25.67&20.93\\

\textbf{ST-Sheaf (Ours)}  &  0.396 & \textbf{0.599} & \textbf{0.859} & \textbf{1.110} & \textbf{1.362} & 24.31 & \textbf{30.10} & \textbf{33.24} & \textbf{19.15} & \textbf{30.18} & \textbf{13.81} & \textbf{15.32} & \textbf{24.26} & \textbf{10.32} \\
\bottomrule
\end{tabular}%
}
\vspace{-5mm}
\end{table*}

\subsection{Overall Results:}
\begin{figure}[t]
    \centering
    \includegraphics[width=0.9\linewidth]{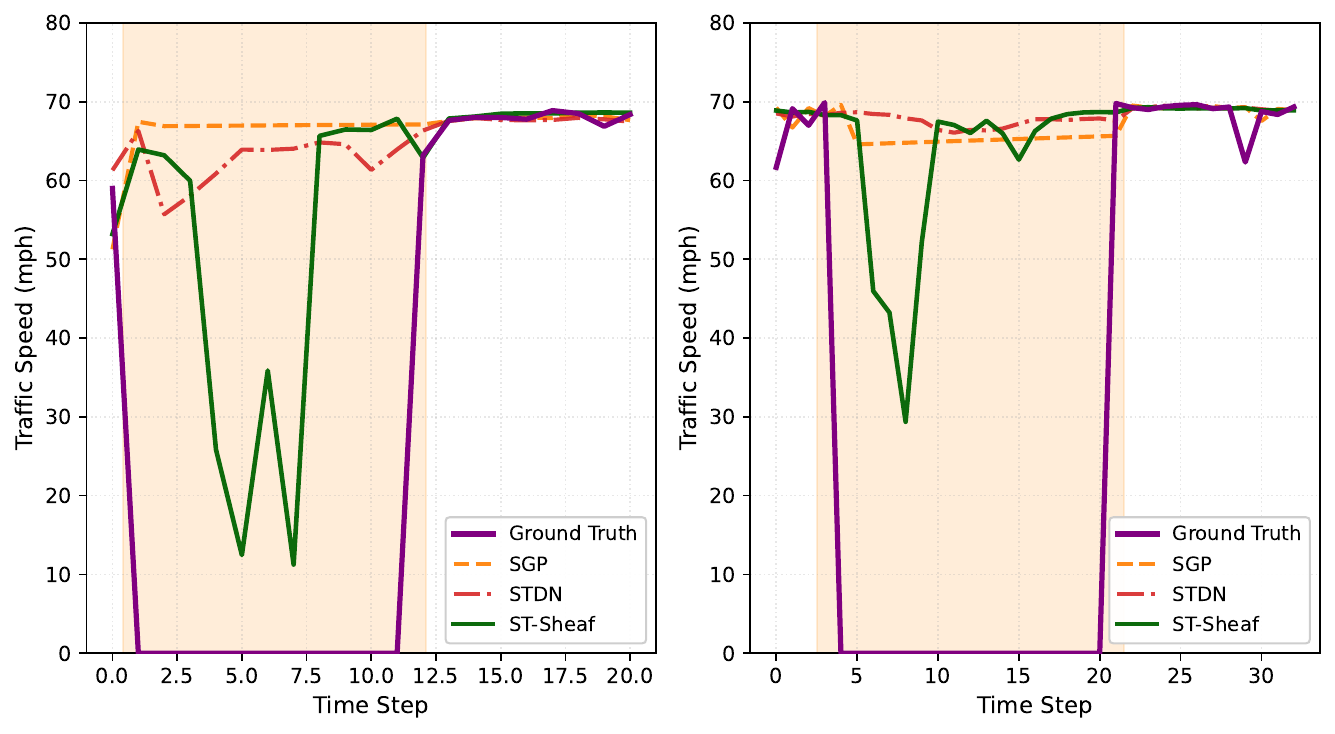}
    \caption{Case study of spatiotemporal dependency modeling on METR-LA test set: A congestion event originating at sensor 36 propagates to affect sensor 62 afterwards. The ST‑Sheaf model captures this cascade effect in the prediction, while baseline methods fail to predict the induced slowdown.}
    \label{fig:pred}
    \vspace{-4pt}
\end{figure}
\begin{figure}[t]
    \centering
    \includegraphics[width=0.88\linewidth]{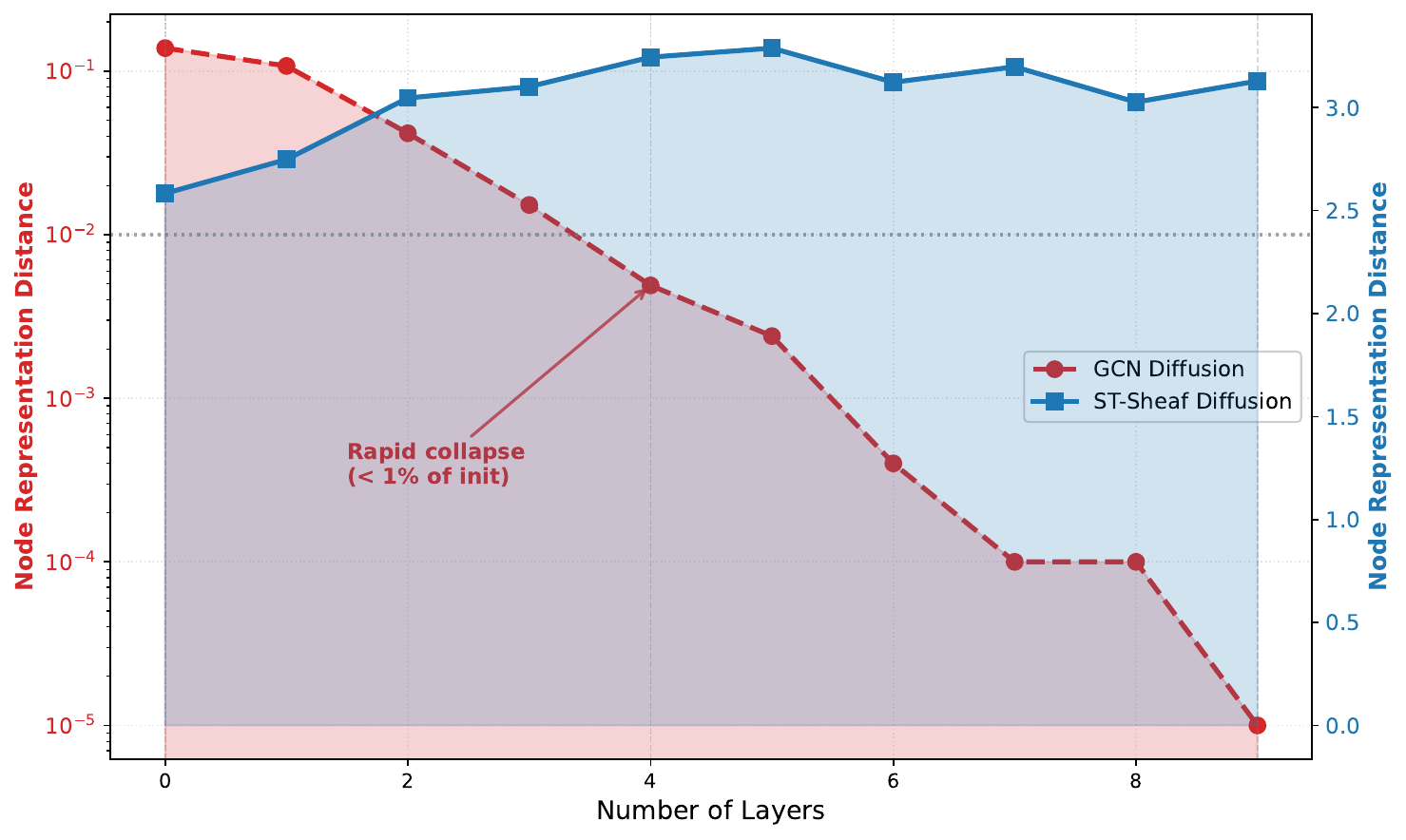}
    \caption{Oversmoothing Analysis: ST-Sheaf GNN diffusion vs. traditional GCN diffusion. The y-axis is the average Euclidean Distance of hidden representation between Connected nodes at each layer.}
    \label{fig:over-smoothing}
    \vspace{-8pt}
\end{figure} 

Tables~\ref{tab:res1} and~\ref{tab:res2} report the full evaluation results across the six benchmarks. Several consistent patterns emerge. Classical sequence models such as ARIMA perform poorly across all settings, as expected given their inability to exploit spatial connectivity and their well-known degradation at longer horizons. GCN-based methods with uniform message passing achieve competitive results on medium-scale datasets but fail to generalize to large-scale, heterogeneous settings such as NAVER-Seoul, where complex and asymmetric local dynamics cannot be adequately captured by fixed propagation rules. ST-Sheaf GNN consistently achieves state-of-the-art or best-in-class performance across all datasets, domains, and forecasting horizons. Most notably, it exhibits the smallest error accumulation from short- to long-term horizons across all benchmarks, as illustrated in Figure~\ref{fig:horizon}, demonstrating that sheaf-based diffusion confers not only higher accuracy but also greater robustness to extended prediction horizons.

Figure~\ref{fig:pred} presents a qualitative case study on the METR-LA dataset, examining a congestion event at sensor 36 characterized by a sharp drop in traffic speed and its downstream propagation to sensor 62. Despite being strong recent models, SGP (AAAI 2023), CITRUS (NeurIPS 2024), and STDN (AAAI 2025) all fail to capture this cascaded dependency, producing predictions that miss the induced slowdown entirely. ST-Sheaf GNN accurately tracks the propagation of the congestion event, demonstrating its capacity to model long-range spatio-temporal dependencies. This qualitative behavior underscores the practical value of dynamic sheaf restriction maps in real-time forecasting of complex networked systems.

\subsection{Oversmoothing Analysis:}
Figure~\ref{fig:over-smoothing} quantifies the oversmoothing behavior of standard GCN diffusion against ST-Sheaf GNN by tracking the average Euclidean distance between hidden representations of connected nodes across layers on the Molene dataset. The GCN baseline undergoes a rapid and irreversible collapse in representation diversity, with inter-node distances falling below 1\% of their initial values after only a few diffusion steps, at which point node representations become indistinguishable. This confirms the well-known failure mode of uniform message passing in deep architectures: by enforcing global representational consistency, repeated aggregation erases the local heterogeneity required to discriminate between nodes. ST-Sheaf GNN, by contrast, maintains stable and well-separated node representations across all 10 layers, preserving discriminative information throughout the network depth. This behavior is a direct consequence of the sheaf diffusion mechanism, which enforces local consistency through edge-specific restriction maps rather than global uniformity. These results provide strong empirical support for our theoretical analysis in Appendix~\ref{app:proof} and confirm that sheaf-based diffusion is an effective and principled solution to oversmoothing in deep spatio-temporal GNNs.

\subsection{Efficiency:} 
\vspace{-2pt}
Table ~\ref{tab:effeciency} further highlights a key advantage of ST-Sheaf GNN: parameter efficiency without sacrificing expressivity. While STDN contains nearly 6 million parameters, ST-Sheaf GNN achieves superior forecasting performance with only $40-59$k parameters, corresponding to a reduction of approximately $150\times$. Additionally, this efficiency does not compromise modeling capacity: ST-Sheaf GNN's parameters scale gracefully with graph size, increasing from 39.6K parameters at 170 nodes to 58.9K at 774 nodes. In contrast, STDN maintains a fixed parameter count of $\sim$6M regardless of graph size, a clear indication of overparameterization that fails to adapt to problem scale. In terms of training efficiency, ST-Sheaf trains 4.0–$5.6\times$ faster per epoch than STDN and 2.8-$4.0\times$ faster than CITRUS across all datasets. 

\begin{table*}[ht!]
\centering
\small
\caption{Number of model parameters, model scalability with respect to graph size $N$, and training time per epoch for different datasets compared to the most state-of-the-art methods: CITRUS (NeurIPS 2024), STDN (AAAI 2025).}
\label{tab:effeciency}
\resizebox{0.9\textwidth}{!}{%
\begin{tabular}{l*{3}{cc}}
\toprule
\multirow{2}{*}{\textbf{Dataset (Graph Size)}} &  \multicolumn{2}{c}{\textbf{CITRUS}} & \multicolumn{2}{c}{\textbf{STDN}} & \multicolumn{2}{c}{\textbf{ST-Sheaf}} \\
 & Time (s) & \#Parameters & Time (s)& \#Parameters & Time (s)&\#Parameters \\
\midrule
PEMS08 (N=170) & 114.0 & 85,951 & 240.1 & 5,876,387 & \textcolor{mygreen}{28.5$\downarrow$} & \textcolor{mygreen}{39,554$\downarrow$} \\
METR-LA (N=207)  & 114.1 &  86,743 & 419.6 & 5,971,107 & \textcolor{mygreen}{40.9$\downarrow$} & \textcolor{mygreen}{40,738$\downarrow$} \\
PEMS04 (N=307)  & 116.2 & 90,335 & 649.3 & 6,227,107 &  \textcolor{mygreen}{53.2$\downarrow$}& \textcolor{mygreen}{43,938$\downarrow$} \\
NAVER-Seoul (N=774) & 231.7 & 105,279 & 833.1 & 7,422,627 & \textcolor{mygreen}{149.2$\downarrow$} & \textcolor{mygreen}{58,882$\downarrow$} \\
\bottomrule
\end{tabular}%
}
\end{table*}

\begin{table}[t]
\centering
\caption{Effect of stalk dimension (d) on performance and efficiency on Molene dataset, horizon 5,
        temporal heads=4,
        sheaf layers=2.}
\label{tab:d_ablation}
\resizebox{0.98\linewidth}{!}{%
\begin{tabular}{lccc}
\toprule
\textbf{Stalk Dim} & \textbf{Epoch Time (s)} & \textbf{\#Parameters} & \textbf{MAE} \\
\midrule
d=1 & 0.358 & 15490 & 1.533\\
d=4 & 0.369 & 17002 & 1.418\\
d=8 &  0.392 & 19410 & 1.378\\
d=16 & 0.453 & 25570 & 1.370\\
d=32 &  0.608 & 43266 & 1.362\\
d=64 & 0.864 & 100162 & 1.361\\
d=128 & 1.625 & 299970 & 1.360\\
\bottomrule
\end{tabular}%
}
\end{table}

\begin{table}[t]
\centering
\caption{Ablation Studies on the effect of each of the model components on different datasets in MAE. For Molene, we select (step-1, 3, 5 as H1, H2, H3).}
\label{tab:ablations}
\resizebox{0.99\columnwidth}{!}{
\begin{tabular}{l@{\hspace{5pt}}l@{\hspace{5pt}}ccc}
\toprule
\textbf{Dataset} &\textbf{Component} & \textbf{H1} & \textbf{H2} & \textbf{H3} \\
\midrule
\multirow{4}{*}{\textbf{METR-LA}} & Static restriction Maps & 2.75& 3.04& 3.38 \\
&W/o Sheaf Diffusion & 3.05 & 3.57 & 4.44\\
&W/o Temporal Encoding & 2.61 & 2.98 & 3.30\\
&\textbf{ST-Sheaf (Full Model) }&\textbf{ 2.56 }& \textbf{2.96} & \textbf{3.25 }\\
\midrule
\multirow{4}{*}{\textbf{Molene}} & Static restriction Maps & 0.44& 0.92& 1.46 \\
&W/o Sheaf Diffusion & 0.56&1.13 & 1.62\\
&W/o Temporal Encoding & 0.41& 0.88 & 1.39\\
&\textbf{ST-Sheaf (Full Model) }& \textbf{0.39} & \textbf{0.85 }& \textbf{1.36} \\
\midrule
\multirow{4}{*}{\textbf{AirQuality}} & Static restriction Maps& 24.65& 31.73& 34.80\\
&W/o Sheaf Diffusion &27.88 &34.21 & 36.72 \\
&W/o Temporal Encoding & 24.45& 30.14& 33.60\\
&\textbf{ST-Sheaf (Full Model)} & \textbf{24.31} & \textbf{30.10} & \textbf{33.24}\\
\bottomrule
\end{tabular}
}
\end{table}

\subsection{Ablation Studies:}
We conduct comprehensive ablation studies to isolate the contribution of each model component. Table~\ref{tab:d_ablation} examines the effect of the sheaf stalk dimension $d$ on performance and computational cost. Increasing $d$ consistently improves forecasting accuracy, confirming that higher-dimensional stalks yield richer local representations and greater model expressivity. However, performance gains diminish beyond $d=32$, where the marginal improvement no longer justifies the additional parameter count and training overhead. Based on this analysis, we set $d \in \{16, 32\}$ across all experiments, selecting the smaller value for datasets with fewer spatial units.

Table~\ref{tab:ablations} reports the performance of four model variants designed to isolate the contribution of each architectural component. First, replacing dynamic restriction maps with static maps derived solely from graph topology (discarding node-state conditioning) leads to consistent performance degradation across all datasets, confirming that spatio-temporal edge semantics are inherently time-varying and cannot be adequately captured by input-independent maps. Second, removing the sheaf structure entirely and substituting standard GCN message passing reduces the model to a temporal encoder coupled with conventional graph convolutions; this variant suffers substantial performance degradation, on par with classic GCN-based baselines, underscoring the critical role of sheaf diffusion in capturing heterogeneous spatial dependencies. Third, retaining the sheaf topology and diffusion while removing the temporal encoding module produces a modest but consistent drop in performance, yet the model still outperforms most baselines.

\section{Conclusion}
\leavevmode\\
We presented ST-Sheaf GNN, a spatio-temporal forecasting framework grounded in cellular sheaf theory that replaces uniform message passing with learned, dynamic restriction maps conditioned on local node states, while provably preventing oversmoothing by admitting heterogeneous node representations. Extensive experiments across six real-world benchmarks spanning traffic, weather, and air quality domains demonstrate consistent state-of-the-art performance, with particular gains on large-scale and long-horizon settings where local heterogeneity is most pronounced. Beyond accuracy, ST-Sheaf GNN achieves this with a fraction of the parameters and training time required by recent baselines, establishing sheaf-theoretic diffusion as both a principled and practical foundation for spatio-temporal graph learning.

\bibliographystyle{siamplain}
\bibliography{example_references}

\newpage

\appendix

\section{Reproducibility Statement} 
\leavevmode\\
We are committed to ensuring the reproducibility of our results. Detailed experimental setup and description of the datasets and data preprocessing steps are provided in Section 4. Pseudo-code of the ST-Sheaf GNN algorithm is provided in Section \ref{sec:algorithm}. The full code with appropriate documentation is available at: \url{https://anonymous.4open.science/r/ST-SheafGNN-6523/}.

\section{ST-Sheaf GNN Pseudo-code}
\leavevmode\\
Algorithm 1 presents the complete forward pass of ST-Sheaf GNN, from spatio-temporal input encoding through sheaf diffusion to final prediction.
\label{sec:algorithm}
\begin{algorithm}[h!]
\caption{ST-Sheaf GNN}
\DontPrintSemicolon
\KwIn{Graph Adjacency Matrix $\mathbf{A} \in \mathbb{R}^{N \times N}$, Input $\mathbf{X} \in \mathbb{R}^{B \times T \times N \times F_{\mathrm{in}}}$, Forecasting Horizon $H$}
\KwOut{Predictions $\hat{\mathbf{Y}} \in \mathbb{R}^{B \times H \times N \times F_{\mathrm{out}}}$}
$\eta_e \gets \big((\deg(s_e)+\varepsilon)(\deg(d_e)+\varepsilon)\big)^{-1/2}$\;
$\mathbf{Z} \gets \mathbf{W}_{\mathrm{temp}} \mathbf{X} \in \mathbb{R}^{B \times T \times N \times D}$\;
\For{$v \gets 1$ \KwTo $N$}{
    $\mathbf{Z}_{:, :, v} \gets \mathrm{MHSA}(\mathbf{Z}_{:, :, v}) $\;
    $\mathbf{Z}_{:, :, v} \gets \mathrm{LayerNorm}\big(\mathbf{Z}_{:, :, v}+ \mathbf{X}_{:, :, v}\big)$\;
}
$\mathbf{H} \gets \mathrm{reshape}(\mathbf{Z}, [BT, N, D])$\;
$\mathbf{H} \gets \mathbf{W}_{\mathrm{stalk}} \mathbf{H} \in \mathbb{R}^{(BT) \times N \times d}$\;
Compute restriction maps $\{\boldsymbol{\rho}_u^e, \boldsymbol{\rho}_v^e\}_{e=1}^E \subset \mathbb{R}^{d \times d}$ from $\mathbf{H}$\;
\For{$\ell \gets 1$ \KwTo $L$}{
    \For{$e = (u,v) \in \mathcal{E}$}{
        $\boldsymbol{\delta}_e \gets (\boldsymbol{\rho}_u^e \mathbf{H}_u) - (\boldsymbol{\rho}_v^e \mathbf{H}_v)$\;
        $\boldsymbol{\delta}_e \gets \eta_e \cdot \boldsymbol{\delta}_e$\;
    }
    \For{$v \in \mathcal{V}$}{
        $\mathbf{M}_v \gets \displaystyle\sum_{e=(v,w) \in \mathcal{E}} \boldsymbol{\delta}_e 
                         - \sum_{e=(w,v) \in \mathcal{E}} \boldsymbol{\delta}_e$\;
    }
    $\tilde{\mathbf{H}} \gets \sigma\big(\mathbf{W}_\ell \mathbf{M}\big)$\;
    $\mathbf{g} \gets \sigma\big(\mathbf{W}_g^\ell [\mathbf{H} \,\|\, \tilde{\mathbf{H}}]\big)$\;
    $\mathbf{H} \gets \mathbf{g} \odot \tilde{\mathbf{H}} + (1 - \mathbf{g}) \odot \mathbf{H}$\;
}
$\hat{\mathbf{Y}} \gets \mathbf{W}_{\mathrm{out}} \mathbf{H} \in \mathbb{R}^{(BT) \times N \times F_{\mathrm{out}}}$\;
Reshape $\hat{\mathbf{Y}}$ to $\mathbb{R}^{B \times H \times N \times F_{\mathrm{out}}}$\;
\Return $\hat{\mathbf{Y}}$\;
\end{algorithm}

\section{Datasets Statistics}
\begin{table*}[t]
\centering
\caption{Summary of datasets characteristics.}
\label{tab:dataset_summary}
\resizebox{0.99\textwidth}{!}{%
\begin{tabular}{lcccccc}
\toprule
 & \textbf{METR-LA} & \textbf{NAVER-Seoul} & \textbf{PEMS04} & \textbf{PEMS08} & \textbf{Molene} & \textbf{AirQuality}\\
\midrule
\textbf{Spatial Units} & 207 & 774 & 307 & 170 & 32 & 437\\
\textbf{Timesteps} & 34,272 & 26,208 & 16992 & 17856 & 744 &  8760 \\
\textbf{Sampling Time} & 5 minutes & 5 minutes & 5 minutes & 5 minutes & 1 hour &  1 hour\\
\textbf{Start Time} & 03.2012 & 09.2020 & 01.2018 & 07.2016  & 01.2014 & 05.2014\\
\textbf{End Time} & 06.2012 & 12.2020 & 02.2018 & 08.2016  & 01.2014 & 04.2015\\

\textbf{Region} & Los Angeles & Seoul & Bay Area & San Bernardino & Brest (France) & 43 Chinese cities \\

\bottomrule
\end{tabular}%
}
\end{table*}
\leavevmode\\
Table \ref{tab:dataset_summary} provides an overview of the six real-world benchmark datasets used to evaluate ST-Sheaf GNN. The datasets span multiple domains and geographic regions, ranging from urban traffic networks in Los Angeles, Seoul, and California, to weather monitoring in Brest, France, and air quality measurements across 43 Chinese cities. They vary substantially in scale, from 32 spatial units (Molene) to 774 (NAVER-Seoul), and in temporal resolution, from 5-minute traffic snapshots to hourly environmental readings.

\section{Error Analysis}
\begin{figure*}[t]
    \centering
    \includegraphics[width=0.99\linewidth]{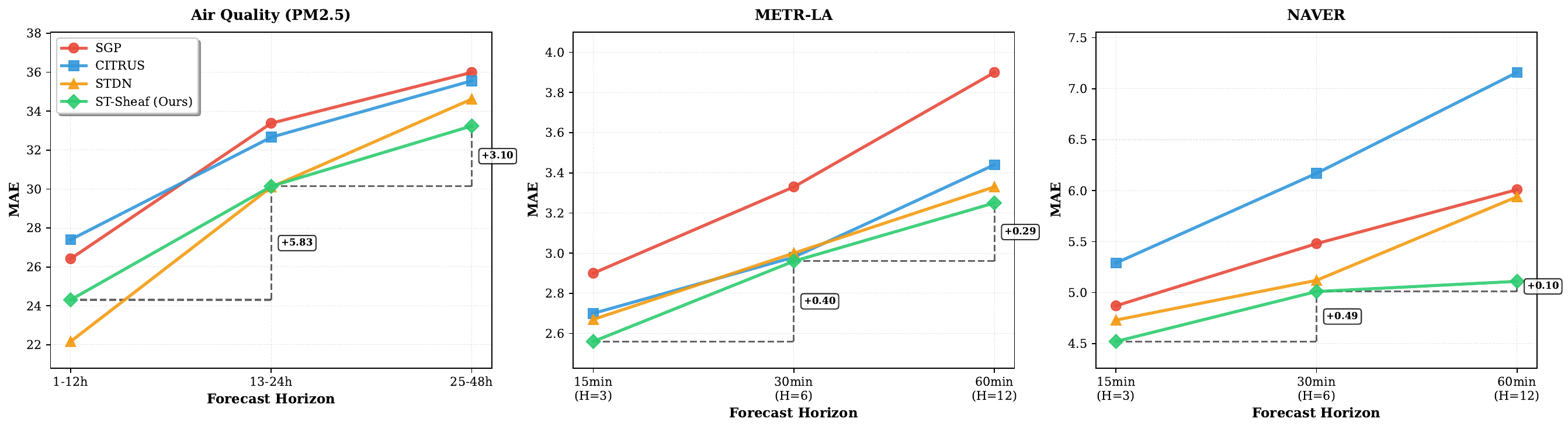}
    \caption{Forecasting error (MAE) across increasing prediction horizons on METR-LA, NAVER-Seoul, and Air Quality datasets. ST-Sheaf exhibits slower error accumulation over longer horizons compared to strong baselines.}

    \label{fig:horizon}
\end{figure*}
\leavevmode\\
Figure \ref{fig:horizon} illustrates how forecasting error (MAE) evolves with increasing prediction horizons across three different datasets: METR-LA, NAVER-Seoul, and Air Quality. While competing methods degrade at longer horizons, ST-Sheaf GNN exhibits the smallest error accumulation. Notably, the margin over the second-best baseline widens at longer horizons — reaching up to 5.83 MAE improvement on Air Quality and 0.49 on NAVER-Seoul at the 60-minute horizon, highlighting the model's ability to maintain predictive accuracy under challenging long-range forecasting conditions.

\section{Theoretical Analysis}
\label{app:proof}
\leavevmode\\
The space of global sections of the sheaf $\mathcal{F}$ over the graph 
$G = (V,E)$ consists of all node-wise assignments that are consistent 
across edges. A global section assigns a vector 
$\mathbf{x}_v \in \mathcal{F}(v)$ to each node $v \in V$, 
such that the corresponding restricted values from adjacent nodes 
agree on every edge $e=(u,v)$. This compatibility condition ensures 
that locally defined data at neighboring nodes remain consistent 
when transported to their shared edge.

To explicitly measure deviations from this consistency condition, 
we define the \emph{coboundary map}
$\delta : C^0(G, \mathcal{F}) \to C^1(G, \mathcal{F})$, 
which maps node-level signals to edge-level discrepancies:
\begin{equation}
(\delta \mathbf{x})_e
=
\rho_{v \trianglelefteq e}\, \mathbf{x}_v
-
\rho_{u \trianglelefteq e}\, \mathbf{x}_u,
\quad
e = (u, v) \in E.
\end{equation}

Under this definition, the space of global sections corresponds 
precisely to the kernel of the coboundary operator:
\begin{equation}
\mathcal{F}(G)
=
\ker(\delta),
\end{equation}
that is, global sections are exactly those node assignments 
whose induced edge discrepancies vanish identically.
The sheaf Laplacian $L_{\mathcal{F}} : C^0(G,\mathcal{F}) \to C^0(G,\mathcal{F})$ is defined as:
\begin{equation}
    L_{\mathcal{F}} = \delta^\top \delta,
\end{equation}
which is symmetric positive semi-definite by construction. For a node $u$, the action of $L_{\mathcal{F}}$ on a signal $\mathbf{h} \in \bigoplus_v \mathcal{F}(v)$ is:
\begin{equation}
    (L_{\mathcal{F}} \mathbf{h})_u = \sum_{e = (u,v) \in E} \rho_{u \trianglelefteq e}^\top \left( \rho_{u \trianglelefteq e}\, \mathbf{h}_u - \rho_{v \trianglelefteq e}\, \mathbf{h}_v \right).
\end{equation}
When $d = 1$ and $\rho_{v \trianglelefteq e} = 1$ for all $(v, e)$, this reduces to the classical graph Laplacian $L = D - A$.

The full sheaf Laplacian $L_{\mathcal{F}} \in \mathbb{R}^{Nd \times Nd}$ has a block decomposition:
\begin{equation}
\begin{aligned}
(L_{\mathcal{F}})_{uu} &= \sum_{e \ni u} 
\rho_{u \trianglelefteq e}^\top \rho_{u \trianglelefteq e}, \\
(L_{\mathcal{F}})_{uv} &= -\rho_{u \trianglelefteq e}^\top 
\rho_{v \trianglelefteq e},\quad e=(u,v).
\end{aligned}
\end{equation}

In ST-Sheaf GNN, each restriction map is parameterized as element-wise multiplication by a learned vector $\mathbf{r}_{u \trianglelefteq e} \in \mathbb{R}^d$:
\begin{equation}
    \rho_{u \trianglelefteq e}\, \mathbf{h}_u = \mathbf{r}_{u \trianglelefteq e} \odot \mathbf{h}_u,
\end{equation}
corresponding to a diagonal matrix $\mathrm{diag}(\mathbf{r}_{u \trianglelefteq e}) \in \mathbb{R}^{d \times d}$. The resulting diagonal sheaf Laplacian block becomes:
\begin{equation}
\begin{aligned}
    (L_{\mathcal{F}})_{uu} = \sum_{e \ni u} \mathrm{diag}(\mathbf{r}_{u \trianglelefteq e})^2, \\
    (L_{\mathcal{F}})_{uv} = -\mathrm{diag}(\mathbf{r}_{u \trianglelefteq e}) \cdot \mathrm{diag}(\mathbf{r}_{v \trianglelefteq e}).
\end{aligned}
\end{equation}
This reduces the parameter count per edge from $\mathcal{O}(d^2)$ to $\mathcal{O}(d)$, and the total parameter cost of the sheaf structure from $\mathcal{O}(|E|d^2)$ to $\mathcal{O}(|E|d)$.

Sheaf diffusion corresponds to gradient flow on the quadratic energy:
\begin{equation}
    \mathcal{E}_{\mathcal{F}}(\mathbf{h}) = \frac{1}{2} \mathbf{h}^\top L_{\mathcal{F}} \mathbf{h} = \frac{1}{2} \sum_{e=(u,v)\in E} \left\| \rho_{u \trianglelefteq e}\, \mathbf{h}_u - \rho_{v \trianglelefteq e}\, \mathbf{h}_v \right\|_2^2.
\end{equation}
The continuous gradient flow equation is:
\begin{equation}
    \frac{d\mathbf{h}}{dt} = -L_{\mathcal{F}}\, \mathbf{h},
\end{equation}
with solution $\mathbf{h}(t) = e^{-t L_{\mathcal{F}}} \mathbf{h}(0)$. Discretizing with step size $\alpha$ yields the diffusion update:
\begin{equation}
    \mathbf{h}^{(\ell+1)} = \mathbf{h}^{(\ell)} - \alpha\, L_{\mathcal{F}}\, \mathbf{h}^{(\ell)} = \left(I - \alpha\, L_{\mathcal{F}}\right) \mathbf{h}^{(\ell)}.
\end{equation}
To guarantee contractive convergence on the non-trivial subspace, the step size $\alpha$ must satisfy:
\begin{equation}
    0 < \alpha < \frac{2}{\lambda_{\max}(L_{\mathcal{F}})},
\end{equation}
where $\lambda_{\max}(L_{\mathcal{F}})$ denotes the largest eigenvalue of the sheaf Laplacian. Under this constraint, the spectral radius of the update operator is strictly less than one.

Let $\lambda_{\min}^+ > 0$ be the smallest non-zero eigenvalue of $L_{\mathcal{F}}$. As $\ell \to \infty$, the diffused signal $\mathbf{h}^{(\ell)}$ converges to the projection onto $\ker(L_{\mathcal{F}})$, i.e., the space of global sections $\mathcal{F}(G)$.

For a standard GCN with scalar Laplacian $L$, $\ker(L)$ consists of constant signals over connected components every node converges to the same representation. In contrast, $\ker(L_{\mathcal{F}})$ is spanned by signals satisfying $\rho_{u \trianglelefteq e}\,\mathbf{h}_u = \rho_{v \trianglelefteq e}\,\mathbf{h}_v$ for all $e=(u,v)$. The theoretical advantage over standard GCNs is that the kernel dimension is scaled by $d$, allowing up to $d$ times more degrees of freedom in the equilibrium. This mitigates oversmoothing by permitting heterogeneous node representations within the same connected component. The dimension of $\ker(L_{\mathcal{F}})$ satisfies:
\begin{equation}
    \dim \ker(L_{\mathcal{F}}) = Nd - \mathrm{rank}(L_{\mathcal{F}}).
\end{equation}
Together, these theoretical results establish that ST-Sheaf GNN is provably resistant to oversmoothing, with a guaranteed minimum representation diversity that grows with both the stalk dimension d and the number of connected components.
\end{document}